\documentclass[10pt,twocolumn,letterpaper]{article}

\usepackage{cvpr}              

\usepackage{multirow}
\usepackage{float}
\usepackage{makecell}
\usepackage{bbding}
\usepackage{pifont}
\usepackage[accsupp]{axessibility}

\definecolor{cvprblue}{rgb}{0.21,0.49,0.74}
\usepackage[pagebackref,breaklinks,colorlinks,allcolors=cvprblue]{hyperref}

\title{MambaVO: Deep Visual Odometry Based on Sequential Matching Refinement and Training Smoothing}

\author{
Shuo Wang$^{1\ast}$,
Wanting Li$^{1\ast}$,
Yongcai Wang$^{1\dagger}$,
Zhaoxin Fan$^{2,3\dagger}$,\\
Zhe Huang$^{1}$,
Xudong Cai$^{1}$,
Jian Zhao$^{4,5}$,
Deying Li$^{1}$\\
[2mm]
$^1$School of Information, Renmin University of China \quad \\
$^2$Beijing Advanced Innovation Center for Future Blockchain and Privacy Computing, \\ Institute of Artificial Intelligence, Beihang University  \quad \\
$^3$Hangzhou International Innovation Institute, Beihang University \quad \\
$^4$The Institute of AI (TeleAI), China Telecom\\
$^5$School of Artificial Intelligence, Optics and Electronics (iOPEN), \quad \\ Northwestern Polytechnical University (NWPU), Xi'an China \\
\normalsize{
$^\ast$Equal contribution} \quad
\normalsize{
$^\dagger$Corresponding authors}
}

\begin{document}
\maketitle

\begin{abstract}

Deep visual odometry has demonstrated great advancements by learning-to-optimize technology. 
This approach heavily relies on the visual matching across frames. However, ambiguous matching in challenging scenarios leads to significant errors in geometric modeling and bundle adjustment optimization, which undermines the accuracy and robustness of pose estimation.
To address this challenge, this paper proposes MambaVO, which conducts robust initialization, Mamba-based sequential matching refinement, and smoothed training to enhance the matching quality and improve the pose estimation.
Specifically, the new frame is matched with the closest keyframe in the maintained Point-Frame Graph (PFG) via the semi-dense based Geometric Initialization Module (GIM). 
Then the initialized PFG is processed by a proposed Geometric Mamba Module (GMM), which exploits the matching features to refine the overall inter-frame matching. 
The refined PFG is finally processed by differentiable BA to optimize the poses and the map. 
To deal with the gradient variance, a Trending-Aware Penalty (TAP) is proposed to smooth training and enhance convergence and stability.  A loop closure module is finally applied to enable MambaVO++. 
On public benchmarks, MambaVO and MambaVO++ demonstrate SOTA performance, while ensuring real-time running.
\end{abstract}
    
\section{Introduction}
\label{sec:intro}

Visual Odometry (VO) is the task of tracking an agent's six-DOF poses when interacting with the physical world, which is critical for robots, self-driving cars and other autonomous agents \cite{huang2024roco, wang2023distributed, wang2023communication}. 
Early works mainly rely on geometric feature methods \cite{mur2015orb,campos2021orb,qin2018vins}, in which, the frontend extracts feature points and conducts feature matching \cite{rublee2011orb,li2023colslam,yang2016pop}, while the backend optimizes poses and build maps using filtering \cite{geneva2020openvins} or  optimization methods \cite{kummerle2011g,dellaert2012factor, wang2024gslamot}. 
However, the geometric features are less effective in challenging environments, such as scenes with weak textures. 

Building on the progress in computer vision and deep learning, early approaches have explored using neural networks to directly regress poses \cite{wang2017deepvo,mohanty2016deepvo, li2018undeepvo, shen2023dytanvo}.
However, these methods suffer from poor accuracy and lack generalization to unseen environments.
A more promising paradigm is the concept of \textit{learning to optimize} \cite{tang2018ba,teed2021droid,teed2024deep}, where neural networks are utilized to extract features and perform matching, while differentiable nonlinear optimization layers \cite{tang2018ba, pineda2022theseus} refine poses based on geometric residuals.
This type of method has achieved state-of-the-art results in the field of deep visual odometry\cite{teed2024deep,lipson2024deep}.

The success of learning-to-optimize deep VO lies on the nested optimization of the sequential image matching network and the differentiable Bundle Adjustment (BA). 
Although being successful, we find that the accuracy and robustness of matching and pose optimization in existing methods are limited by three key aspects.
(1) \textbf{Unstable initialization}. Existing methods
using random patches \cite{teed2024deep}, predicted optical flow \cite{teed2021droid}, or motion prediction \cite{keetha2024splatam} may fail to provide robust initial estimates in challenging environments. 
(2) \textbf{Less refined matching}. Accurate inter-frame correspondences are crucial for visual odometry. Current feature extraction and limited feature interaction restrict the matching precision \cite{teed2024deep,lipson2024deep}. 
(3) \textbf{The training challenges}: The gradient variance problem is a main challenge for the nested optimization in learning to optimize, which causes instability and slow convergence \cite{gurumurthy2024variance}. While some methods adjust the loss via gradient weighting \cite{gurumurthy2024variance}, this can lead to overfitting, limiting generalization to unseen environments.

To tackle the issue, we propose MambaVO, a novel VO system that refines inter-frame image matching and produces more accurate and robust poses using a Mamba-based module.
MambaVO introduces three key components. First, a Point-Frame Graph (PFG) captures observation relationships, enabling robust initialization and sequence-based matching refinement. Second, the Geometric Initialization Module (GIM) uses semi-dense matching \cite{wang2024efficient} and PnP \cite{PoseLib,zheng2013revisiting} to predict pixel correspondences and initialize poses, extracting features for fusion in subsequent matching refinement. Third, the Geometric Mamba Module (GMM) refines pixel matches by leveraging historical tokens and fusion features through Mamba blocks, adjusting pixel correspondences and frame weights within the PFG. To ensure stable training, we further introduce the Trending-Aware Penalty (TAP), which balances pose and matching losses, addressing gradient variance issues \cite{gurumurthy2024variance}. We also extend MambaVO to MambaVO++, incorporating loop closure for global optimization in SLAM. 

Our contributions are summarized as follows:
\begin{itemize}
\item We propose a novel VO system, MambaVO, designed to address Unstable initialization, less refined matching, and training challenges in deep VO methods.
\item We introduce three key components to enhance MambaVO: (1) the Geometric Initialization Module (GIM), which ensures robust pose initialization using semi-dense matching and PnP; (2) the Geometric Mamba Module (GMM), which refines pixel correspondences using historical tokens and fusion features; and (3) the Trending-Aware Penalty (TAP), which stabilizes training by balancing pose and matching losses.
\item We extend the system to MambaVO++, incorporating loop closure for global optimization in a full SLAM system.
Through extensive experiments on EuRoC \cite{burri2016euroc}, TUM-RGBD \cite{sturm12iros}, KITTI \cite{geiger2012we}, and TartanAir \cite{wang2020tartanair}, MambaVO and MambaVO++ achieve state-of-the-art performance in accuracy, while ensuring real-time performance and lower GPU memory consumption compared to recent learning-to-optimize methods.
\end{itemize}
\color{black}

\section{Related Work}
\label{sec:relatedwork}

\begin{figure*}[ht]
  \centering
  \includegraphics[width=0.98\linewidth]{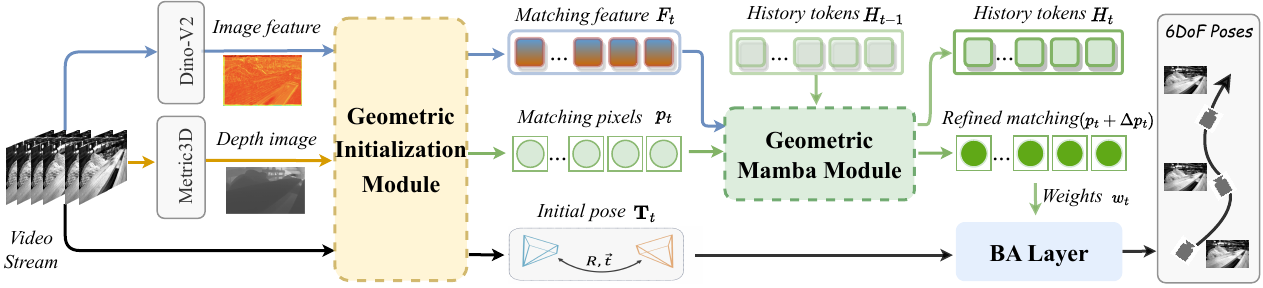}
  \caption{The proposed MambaVO extracts Dino-v2\cite{oquab2023dinov2} features from the input RGB sequence and estimates the depth\cite{hu2024metric3d} for keyframes. In the Geometric Initialization Module (\cref{sec:initialization}), a semi-dense matching network is utilized to generate initial matches, estimate the initial poses, and extract features for each match. Next, the Geometric Mamba Module(\cref{sec:GMM}) refines and re-weights the matching. Finally, we use a differentiable bundle adjustment (BA) to optimize the final poses, ensuring accuracy and stability in the pose estimation process.}
  \label{fig:pipeline}
\end{figure*}
Traditional visual odometry and SLAM methods have produced numerous notable approaches, including direct methods like DSO \cite{gao2018ldso}, SVO \cite{forster2014svo}, and LSD-SLAM \cite{engel2014lsd}, as well as feature-based methods such as ORB-SLAM \cite{mur2015orb}, VINS-Mono \cite{qin2018vins}, and OpenVINS \cite{geneva2020openvins}. However, we focus on deep visual odometry, which forms the core of our study.
\subsection{Direct Pose Regression for Visual Odometry}
Early deep visual odometry methods, which we refer to as direct pose regression approaches, applied neural networks to predict relative poses from visual inputs \cite{mohanty2016deepvo,wang2017deepvo,clark2017vinet}. 
DeepVO \cite{mohanty2016deepvo} was one of the first works to use CNNs to predict rotation and translation changes from sequential images. 
VINet \cite{clark2017vinet} and VidLoc \cite{clark2017vidloc} framed visual-inertial odometry as a sequence-to-sequence learning problem. In unsupervised learning, methods aimed to estimate depth and pose without labeled data. SfM-Learner \cite{zhou2017sfm-learner} used photometric consistency to predict both pose and depth. Although it outperformed traditional methods like ORB-SLAM \cite{mur2015orb} (without loop closure), it still struggled with generalization. SfM-Net \cite{2017sfm-net} took this further by jointly estimating optical flow, scene flow, and 3D point clouds. UnDeepVO \cite{li2018undeepvo} used stereo images during training to address the scale ambiguity in monocular VO, improving accuracy.

While these methods demonstrated promising results, they often exhibited limitations in accuracy and generalization across diverse environments. Therefore, this paper explores a more stable learning-to-optimize approach.

\subsection{Learning to Optimize for Visual Odometry}
The Learning to Optimize approach has gained prominence with the adoption of differentiable bundle adjustment \cite{tang2018ba, pineda2022theseus}. Early works, such as BA-Net \cite{tang2018ba}, introduced a bundle adjustment layer to optimize photometric re-projection error. Some other methods refined pose estimates via geometric constraints predicted by neural networks \cite{clark2018learning, jatavallabhula2020slam} and learned correspondence weights to improve optimization \cite{muhle2023learning, ranftl2018deep}. 
DROID-SLAM \cite{teed2021droid} improved upon this by refining per-pixel depth using advanced optical flow, overcoming the limitations of basis depth maps, and employing dense bundle adjustment to further enhance accuracy. DPVO \cite{teed2024deep} introduced patch tracking with refinement, leading to more efficient pose estimation while reducing computational costs. Its follow-up work, DPV-SLAM \cite{lipson2024deep}, extended global optimization by leveraging a patch-graph structure. More recently, V2V \cite{gurumurthy2024variance} identified training challenges caused by gradient variances and mitigated the issue by dynamically weighting the loss. 

While these approaches have advanced visual odometry performance, they suffer from less refined matching and inaccuracy poses in low-texture or repetitive environments. In this paper, we propose MambaVO to tackle the issues.

\subsection{State Space Models}
State Space Models (SSMs) \cite{gu2021efficiently, gu2022parameterization} have garnered significant attention due to their ability to capture long-range dependencies with linear complexity, making them highly effective across a variety of fields. A recent advancement in SSMs, the Mamba architecture \cite{gu2023mamba}, introduces a selection mechanism to extract features from sequential data more efficiently. 
Mamba has achieved outstanding results in diverse applications. Vision Mamba \cite{zhu2024vision} enhances visual representation by scanning the visual space for better feature extraction. VideoMamba \cite{li2024videomamba} efficiently processes video patches to learn temporal information. In collaborative perception, Mamba fuses spatial features from multiple agents \cite{li2024collamamba}, highlighting its versatility in multi-agent systems.

In our work, we model VO as a task of refining image matching and poses across sequences. Leveraging Mamba blocks with semi-dense initialization, we enhance the ability to fuse and refine matching across frames, thereby significantly improving accuracy and robustness. To the best of our knowledge, this is the first time SSMs have been employed in the context of visual odometry. 
\section{Methods}
The overview of MambaVO is shown in \cref{fig:pipeline}.
Given a sequence of  RGB images as input, MambaVO solves the set of camera poses $\{\mathbf{T}_\tau\}_{\tau=0}^N$ and the reconstructed map points $\{\mathrm{P}_i\}_{i=0}^{L}$,
where $\mathbf{T_\tau} \in SE(3)$ represents the camera pose of frame $\tau$, and $\mathrm{P_i}=(x_i,y_i,z_i)^\top$ denotes the position of map point $i$ in the world coordinate system. 

We model the observation relationship between the map points and the cameras, as well as the frame co-visibility, using a \emph{Point-Frame Graph (PFG)} $\mathcal{G}=(\mathcal{V}, \mathcal{E})$.
The PFG vertices represent camera poses and map points.
The edge between camera frames (frame-frame edge), such as $(t-1,t)$ represents the pose transformation $\mathbf{T}_{t-1,t}$ between them.
Frame-point edges represent projections. An edge $(i, t) \in \mathcal{E}$ indicates the camera observes the map point $i$ in the $t$-th frame. We define the observed pixel coordinates as $\mathrm p_t^i = (u_t^i, v_t^i)^\top$. 
The projection relationship for edge $(i, t)$ is described using the projection function: 
\begin{equation}
	\mathrm p_t^i = \Pi(\mathbf{T_t}, \mathrm P_i) = \mathbf{K} \mathbf{T_t} \mathrm P_i
\vspace{-0.2cm}
\end{equation}
where $\mathbf K$ is the intrinsic matrix of the camera.
If both edges $(i, t-1)$ and $(i, t)$ exist, it means that the same map point is observed in both the $t-1$ frame and the $t$ frame. This co-visibility and association is initialized by image matching in the initialization step (\cref{sec:initialization}).
To minimize computational costs, we maintain only the latest PFG within a sliding window containing $\mathcal W$ frames. The PFG is initialized from the first frame pair.

\subsection{Geometric Initialization Module: GIM}
\label{sec:initialization}
\begin{figure}[t]
  \centering
  \includegraphics[width=0.9\linewidth]{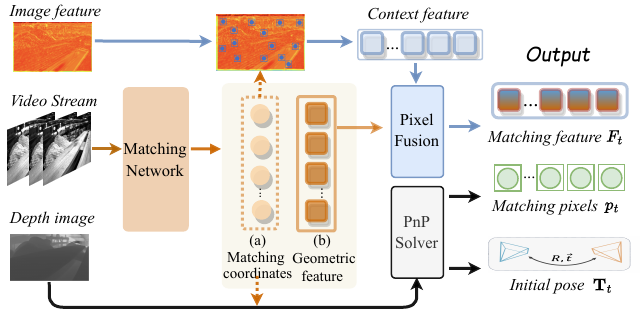}
  \caption{Illustration of Geometric Initialization Module. GIM extracts geometric and context features, and performs matching using a semi-dense matching network. The initial pose is estimated using a PnP solver based on the matched pixels.}
  \label{fig:init}
\end{figure}
\subsubsection{Matching and Pose Initialization.}
After receiving the $t$-th image, the nearest keyframe $r$ in time order in PFG serves as the reference frame for initial matching and feature extraction.
We utilize Dino-v2 \cite{oquab2023dinov2} to extract context features, and EfficientLoFTR \cite{wang2024efficient} provides $k$ initially matched coordinates $\{[p_r^i,p_t^i]\}_{i=1}^{k}$ and pixel-wise geometric features $\mathbf{G}_t=\{\mathbf{G}_t^i\}_{i=1}^{k}$ from the fine-matching module in EfficientLoFTR. 
These $k$ matches are used to estimate the initial pose $\mathbf T_t$ of the frame $t$.  We apply Metric3D \cite{hu2024metric3d} to estimate the  depth for each matched point in the reference frame, denoted by $\{d_{r}^i\}_{i=1}^{k}$. By obtaining the 3D points and by the 3D-2D correspondences,  PnP (Perspective-n-Point) \cite{PoseLib} calculates the initial pose $\mathbf T_t $ and refine the 3D coordinates of the matched points:
\begin{align}
    \mathbf T_{r,t}, \{\mathrm{P}_i\}_{i=1}^{k} &= \text{PnP}(\{[p_r^i,p_t^i]\}_{i=1}^{k} , \{d_r^i\}_{i=1}^{k})  \\
    \mathbf T_t &= \mathbf T_{r,t}  \mathbf T_{r} 
\end{align}
where $\mathbf T_{r,t}$ is the relative pose between frame $r$ and $t$.

\subsubsection{Matching Feature Preparation}
As in \cref{fig:init}, we prepare matching features for matching refinement by the Geometric Mamba Module (\cref{sec:GMM}). 
The matched pixels in frame $i$ are used to query the context features from Divo-v2 encoder, yielding the pixel-wise context features $\mathbf{C}_t=\{\mathbf{C}_t^i\}_{i=1}^{k}$.
Context features $\mathbf{C}_t$ and geometric features $\mathbf{G}_t$ are fused via concatenation, 1D convolution, and ReLU activation to obtain the fused matching feature $\mathbf{F}_t^i$ for each matched pixel in the pixel fusion process:
\begin{equation}
    \mathbf{F}_t^i = \text{ReLU}(\text{1DConv}(\text{Concat}(\mathbf{G}_t^i, \mathbf{C}_t^i))), i \in [1, k]
\end{equation}
where $\mathbf{F}_t^i \in\mathbb{R}^{1\times 384}$. $\mathbf{F}_t\in\mathbb{R}^{k\times 384}$ represents all matching features in the current frame $t$.
The initialized PFG will then be processed by the Geometric Mamba Module. 

\subsection{Geometric Mamba Module: GMM}
\label{sec:GMM}
\begin{figure}[ht]
  \centering
  \includegraphics[width=0.95\linewidth]{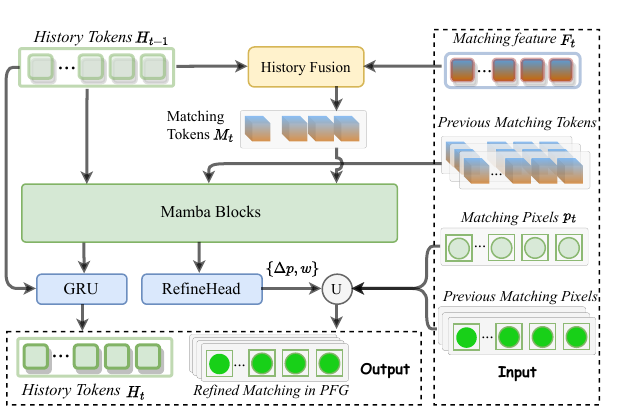}
  \caption{Illustration of Geometric Mamba Module. The GIM refines the matching by incorporating both current and historical information. Matching tokens $\mathbf M_t$ are derived from matching features and previous matching tokens, then processed through Mamba blocks. The RefineHead further decodes the refinement and the weight for each matching.}
  \label{fig:mamba}
\end{figure}

GMM refines the matching coordinates of each frame at the pixel level by sequentially learning the matching features in PFG.
It accepts the PFG as input, including the matching points $\{[p_r^i,p_t^i]\}_{i=1}^{k}$ for each frame pair and the corresponding matching features $\mathbf{F}_t$.
Using these matching features and the history tokens, GMM refines the matching coordinates and estimates the matching weights.
GMM consists of two main parts: History Fusion and Geometric Mamba Blocks. The detailed structure is in \cref{fig:mamba}.

\subsubsection{History Fusion}
Odometry is essentially a time series problem, and historical information affects current inter-frame matching through sequential pose transformation and overlapping areas.   
We maintain the history token $H_{t-1}\in\mathbb{R}^{k\times 384}$ to represent the historical information and fuse with $\mathbf F_t$ in History Fusion.
The history token for the first frame is initialized from the mathing features in GIM.
Specifically, we perform cross-attention between $\mathbf F_t^i$ and $H_{t-1}$ on each edge in PFG:
\begin{equation}
    \hat{M}_t^i = \text{CrossAttention}(\mathbf F_t^i, H_{t-1}), i \in [1, k]
\end{equation}
Then, we concatenate all $ \hat{M}_t^i $ of the current frame and perform linear mapping on the edge dimension to conduct 1D convolution:
\begin{equation}
    \{{M}_t^i\}_{i=1}^{k} = \text{Conv1D}\left(\text{Concat}\left(\{\hat{M}_t^i\}_{i=1}^{k}\right)\right)
\end{equation}
We then repack $\{{M}_t^i\}_{i=1}^{k}$ to $\mathbf M_t \in\mathbb{R}^{k\times 384}$ to represent the \emph{matching tokens} of all the $k$ edges related to the current frame $t$, which are used in the subsequent Mamba blocks.
This matching feature and history combination design can effectively integrate history information, providing stronger consistency and stability for the final feature expression.

\subsubsection{Geometric Mamba Blocks}
Geometric Mamba Blocks contains $B$ vanilla Mamba blocks\cite{gu2023mamba},  receiving the historical token $H_{t-1}$ along with all the matching tokens $\{\mathbf M_\tau\}_{\tau=t-\mathcal W}^{t}$ from PFG, where the number of vertices is $\mathcal W+1$. 
Geometric Mamba Blocks output an updated history token $\hat{H}_t$ and the refined matching tokens $\{\mathbf M_\tau\}_{\tau=t-\mathcal W}^{t}$ for PFG.  
\begin{equation}
	\hat{H}_t, \{\mathbf M_\tau\}_{\tau=t-\mathcal W}^{t} = \text{MambaBlocks}\left(H_{t-1}, \{\mathbf M_\tau\}_{\tau=t-\mathcal W}^{t}\right)
\end{equation}

Each updated matching token $M_\tau^i \in \mathbf M_\tau$ is further decoded by a Matching Refinement Head, i.e. \textrm{RefineHead()}. 
For the edge $(i,\tau)$ in PFG and the corresponding matching token $M_\tau^i$,  the Matching Refinement Head outputs the pixel refinement $\Delta p_\tau^i$ and the matching weights $w_\tau^i$ respectively, which are subsequently used to update the edge $(i,\tau)$ and the pixel coordinates $p_\tau^i$.
\begin{equation}
\begin{aligned}
    \Delta p_\tau^i, w_\tau^i &= \text{RefineHead}\left( M_\tau^i \right) \\
    \tau  \in & [t-\mathcal W, t], i \in [1, k]
\end{aligned}
\end{equation}
where $\Delta p_\tau^i \in \mathbb{R}^2$ and $w_\tau^i \in \mathbb{R}^2$ represent the adjusted  pixels and the weight for the matching. RefineHead() contains two MLPs for decoding $\Delta p_\tau^i$ and $w_\tau^i$ respectively.
The updated token $\hat{H}_t$ is then combined with $H_{t-1}$ from the previous time step through a GRU module to generate the new history token $H_t$.
\begin{equation}
	H_t = \text{GRU}\left(\hat{H}_t, H_{t-1}\right)
\end{equation}
From above process, it can be seen that GMM takes the result of GIM as input and continuously refines and re-weights the matching in all the frame vertices in the PFG.
With the refinement and evaluated weight for each matching, the cameras' poses and map points are updated through the differentiable bundle adjustment layer\cite{teed2021tangent}. 
The predicted $\Delta p_\tau$ and $w_\tau$ are engaged in the objective function. 
\begin{equation}
	\mathbf T, \mathrm P =\mathop{\arg\min}\limits_{\mathbf T, \mathrm P} \sum_{(i,\tau)\in \mathcal{E}} \Vert \Pi(\mathbf T_\tau, \mathrm P_i) - (\mathrm p_\tau^i + \Delta p_\tau^i) \Vert_{w_\tau^i}
\end{equation}
where $\Vert \cdot \Vert_{w_\tau^i}$ is the Mahalanobis distance weighted by ${w_\tau^i}$. 
We employ the Gauss-Newton method in the BA layer\cite{teed2021tangent} to linearize the objective, update the poses and map points while keeping the matching constant, ensuring that the induced trajectory adjustments are consistent with the predicted matching updates.
\subsection{Smoothed Training}
\label{sec:smoothing}
During training, we jointly train GIM and GMM with the BA layer, an implicit nonlinear optimization module. However, nested optimization causes loss oscillations and convergence challenges. To address this, we introduce the Trending-Aware Penalty to stabilize the loss and improve convergence.
\subsubsection{Training Loss}
We apply supervision to the poses and the matching as in \cite{teed2021droid,teed2024deep}.
In training, we randomly select a set of trajectory segments and each works as a training sample, whose camera poses and the matching ground truth are used as the supervision signal. The PFG for an arbitrary training sample is denoted as $\mathcal G=(\mathcal{V}, \mathcal{E})$.
The final predicted poses are compared with the ground truth poses to compute the relative error loss between each two frames in $\mathcal V$:  
\begin{equation}
	\mathcal L_{pose} = \sum_{\tau_1,\tau_2 \in \mathcal{V}} \Vert Log_{SE(3)}\left[ (\tilde{\mathbf T}_{\tau_1}^{-1} \tilde{\mathbf T}_{\tau_2})^{-1} (\mathbf T_{\tau_1}^{-1} \mathbf T_{\tau_2}) \right]  \Vert
\end{equation}
where $\tilde{\mathbf T}$ is the ground truth and $\mathbf T$ is the predicted pose.

We also supervise the estimated matching pixels and the ground truth. The weights from GMM are used to weight the loss, to focus on the matching pixels that are more important for pose estimation. 
\begin{equation}
	\Hat{\mathcal{L}}_{match} = \sum_{(i,\tau) \in \mathcal{E}} \Vert \tilde{p}_\tau^i - p_\tau^i \Vert_{w_\tau^i}
\end{equation}
where $p_\tau^i$ is the predicted matching coordinates, and $\tilde{p}_\tau^i$ is the ground truth.

\subsubsection{Trending-Aware Penalty}
The gradient of the loss exhibits significant fluctuations due to the variance of trajectories (e.g., motion patterns, lighting, textures), as reported in \cite{gurumurthy2024variance,teed2024deep}, and is also seen in our experiments (\cref{fig:loss}). 
Further,  there are differences in the learning speed of the pose loss and the matching loss.
These lead to  negative impacts on the training convergence.

We propose a \textbf{ Trending-Aware Penalty (TAP)} to solve the above training challenges.
We first calculate the \emph{gradient weighting parameter} as in \cite{gurumurthy2024variance}, which evaluates the gradient difference of the pose loss and the matching loss. 
We adjust the matching loss by the gradient weighting parameter every 50 training iterations to balance the pose loss gradient and the matching loss gradient. 
\begin{equation}
	\mathcal L_{match} = \frac{\|\nabla_\theta\mathcal{L}_\mathrm{pose}\|_2}{\|\nabla_\theta{L}_{match}\|_2} \Hat{\mathcal{L}}_{match}
\end{equation}

Considering the variance brought by different trajectories, we further propose a \emph{trend-based balance parameter} by averaging the losses over historical training iterations. For the $t$-th training iteration, $\Lambda_{match}$ evaluates the decreasing trend of matching loss by comparing with the average loss of the past four iterations. 
\begin{equation}
	\Lambda_{match} = \exp{\left(\frac{\mathcal{L}_{match}^{t}}{\frac{1}{4} \sum_{k=0}^{3}\mathcal{L}_{match}^{t-k}}\right)}  
\end{equation}
We get $\Lambda_{pose}$ in the same way and obtain the trend-based balance parameter, i.e., $\Lambda = \frac{\Lambda_{match}}{\Lambda_{pose}}$. 
$\Lambda$ measures the trend difference. 
Finally, the total loss is balanced by the trend difference to emphasize the one that decreases more slowly. 
\begin{equation}
	\mathcal{L}=\mathcal{L}_{pose}+\Lambda\mathcal{L}_{match}
\end{equation}

\section{MambaVO and MambaVO++}
We implement a complete visual odometry system, \textbf{MambaVO}, based on the proposed network modules. The system takes RGB sequence as input and performs real-time camera tracking and map reconstruction. To ensure high efficiency and accuracy, we adopt a keyframe strategy. To correct the accumulated drift, we further develop \textbf{MambaVO++}, which is a SLAM system incorporating loop closure and global optimization to enhance the overall performance.

\textbf{Keyframes and Optimization.}
A frame is selected as a keyframe if it meets either of the two conditions: (1) the parallax with the previous frame exceeds 30px, or (2) none of the last three frames are keyframes. 
We maintain a sliding window of keyframes ($\mathcal{W}$ is set to 10 in our experiments).

\textbf{Loop Closure and Global Optimization.}
We use a DBoW2-based loop closure strategy \cite{campos2021orb}.
We extract ORB features from keyframes and store all key frames to maintain the global pose graph. 
DBoW2 \cite{GalvezTRO12} performs place recognition to retrieve loop closure pairs.
For loop closure pairs, we calculate the relative pose using matching and the PnP in \cref{sec:initialization}. Based on the loop closure edges, we run a global optimization to correct the accumulated drift.  
These processes are completed in parallel on the CPU.
\begin{figure*}[ht]
  \centering
  \includegraphics[width=0.8\linewidth]{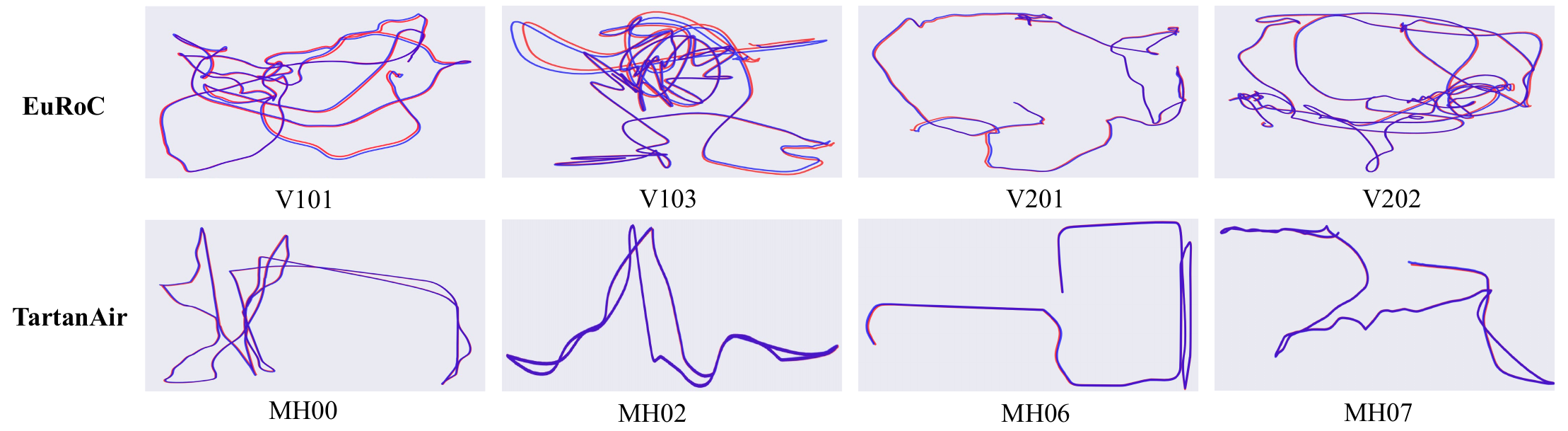}
  \caption{Qualitative visualization. The blue line represents the trajectory estimated by MambaVO, and the red line represents the ground truth. The results show that our estimated trajectory almost completely coincides with the ground truth.}
  \label{fig:viz}
\end{figure*}
\section{Experiments}

We evaluate MambaVO and MambaVO++ on four datasets: EuRoC \cite{burri2016euroc}, TUM-RGBD \cite{sturm12iros}, KITTI \cite{geiger2012we}, and TartanAir \cite{wang2020tartanair}. Our model is trained only on the TartanAir training set, without retraining or fine-tuning on other datasets, which is the same as \cite{teed2024deep,gurumurthy2024variance,teed2021droid}. We train our network for 196k steps with a batch size of 1. Training takes 2.5 days on one RTX 4090 GPU.

\subsection{Analysis on Pose Estimation}
We report pose estimation results on the TartanAir test-split from the CVPR 2020 SLAM competition in  \cref{tab:tartanAir}, EuRoC in  \cref{tab:EuRoC}, KITTI in  \cref{tab:kitti}, and TUM-RGBD in  \cref{tab:tum}. 
We run evaluations five times and report the mean ATE (Absolute Trajectory Error) results from the EVO tool \cite{grupp2017evo} under a monocular setting only. 
We separate methods that without and with loop closure to ensure a fair comparison with MambaVO and MambaVO++ respectively. The qualitative results are shown in \cref{fig:viz}.

As the results in \cref{tab:tartanAir}-\cref{tab:tum}, our method performs robustly across all scenarios, while previous methods fail on certain sequences (marked “\ding{55}”). 
Our Methods achieve the SOTA results (on AVG column) in both loop closure settings. Notably, MambaVO reduces the pose estimation error by 19\%-22\% on average comparing to DROID-VO \cite{teed2021droid}, DPVO \cite{teed2024deep}, and V2V \cite{gurumurthy2024variance} on indoor datasets like EuRoC, TUM-RGBD, and TartanAir.

We can also observe that loop closure and global optimization significantly enhance localization accuracy. In particular, on long sequences like KITTI and EuRoC, MambaVO++ achieves over 50\% improvement compared to MambaVO, demonstrating the benefits of traditional loop closure and global optimization.

\begin{table*}[htbp]
\centering
\scalebox{0.76}{
\begin{tabular}{c|ccccccccc|c}
\toprule
                            & Method          & MH000 & MH001                        & MH002                        & MH003                        & MH004                        & MH005                        & MH006                        & MH007 & AVG                             \\ \midrule
                            & ORBSLAM3\cite{campos2021orb}        & 15.44                        & 2.92                         & 13.51                        & 8.18                         & 2.59                         & 21.91                        & 11.70                         & 25.88                        & 14.38                           \\
                            & COLMAP\cite{schoenberger2016sfm}          & 12.26                        & 13.45                        & 13.45                        & 20.95                        & 24.97                        & 16.79                        & 7.01                         & 7.97                         & 12.5                            \\
                            & DeFlowSLAM\cite{ye2022deflowslam}      & 0.63                         & 0.06                         & \textbf{0.02} & \textbf{0.01} & 2.8                          & 0.20                          & 0.31                         & 0.45                         & 0.56                            \\
                            & DROID-SLAM\cite{teed2021droid}      & \textbf{0.08} & \underline{0.05} & \underline{0.04} & \underline{0.02} & \textbf{0.01} & 0.68                         & 0.30                          & \underline{0.07} & 0.33                            \\
                            & DPV-SLAM\cite{lipson2024deep}        & 0.23                         & \underline{0.05} & \underline{0.04} & 0.04                         & 0.54                         & \underline{0.15} & \underline{0.07} & 0.14                         & \underline{0.16}    \\
\multirow{-6}{*}{\rotatebox{90}{w. loop}}   & MambaVO++(Ours) & \underline{0.12} & \textbf{0.04} & \textbf{0.02} & \underline{0.02} & \underline{0.37} & \textbf{0.14} & \textbf{0.05} & \textbf{0.05} & \textbf{0.10} \\ \midrule
                            & TartanVO\cite{wang2021tartanvo}        & 4.88                         & 0.26                         & 2.00                            & 0.94                         & 1.07                         & 3.19                         & 1.00                            & 2.04                         & 1.92                            \\
                            & DROID-VO\cite{teed2021droid}        & 0.32                         & 0.13                         & 0.08                         & 0.09                         & 1.52                         & 0.69                         & 0.39                         & 0.97                         & 0.58                            \\
                            & DPVO\cite{teed2024deep}            & \underline{0.21} & 0.04                         & \underline{0.04} & \underline{0.08} & \underline{0.58} & \textbf{0.17} & \underline{0.11} & \underline{0.15} & \underline{0.17}    \\
                            & V2V\cite{gurumurthy2024variance}             & 0.18                         & \underline{0.03} & \textbf{0.03} & \textbf{0.02} & \underline{0.58} & 0.30                          & \textbf{0.08} & \textbf{0.05} & 0.18                            \\
\multirow{-5}{*}{\rotatebox{90}{w.o. loop}} & MambaVO (Ours)  & 0.24                         & \textbf{0.02} & \textbf{0.03} & \textbf{0.02} & \textbf{0.46} & \underline{0.18} & 0.13                         & \textbf{0.05} & \textbf{0.14}    \\ \bottomrule
\end{tabular}}
\caption{ATE[m]$\downarrow$ results on the TartanAir test split. The top one is in \textbf{bold} and the second is \underline{underlined}.}
\label{tab:tartanAir}
\end{table*}

\begin{table*}[]
\centering
\scalebox{0.78}{
\begin{tabular}{c|cccccccccccc|c}
\toprule
                            & Method          & MH01    & MH02                          & MH03                          & MH04                          & MH05                          & V101                          & V102                          & V103   & V201                          & V202                          & V203                          & AVG                           \\ \midrule
                            & ORBSLAM3\cite{campos2021orb}        & 0.016                          & 0.027                         & 0.028                         & 0.138                         & 0.072                         & \textbf{0.033} & 0.015                         & 0.033                         & 0.023                         & 0.029                         & \ding{55}                          & -                             \\
                            & LDSO\cite{gao2018ldso}            & 0.046                          & 0.035                         & 0.175                         & 1.954                         & 0.385                         & 0.093                         & 0.085                         & -                             & 0.043                         & 0.405                         & -                             & -                             \\
                            & GO-SLAM\cite{zhang2023go}         & 0.016                          & \textbf{0.014} & 0.023                         & \underline{0.045} & 0.045                         & 0.037                         & \underline{0.011} & 0.023                         & 0.016                         & 0.010                          & 0.022                         & 0.024                         \\
                            & DROID-SLAM\cite{teed2021droid}      & \underline{0.013}  & \textbf{0.014} & \underline{0.022} & \textbf{0.043} & 0.043                         & 0.037                         & 0.012                         & 0.02                          & \underline{0.017} & \underline{0.013} & \textbf{0.014} & \underline{0.022}                         \\
                            & DPV-SLAM\cite{lipson2024deep}        & \underline{0.013}  & \underline{0.016} & \underline{0.022} & \textbf{0.043} & \underline{0.041} & \underline{0.035} & \textbf{0.008} & \textbf{0.015} & 0.020                          & \textbf{0.011} & 0.040                          & 0.024                         \\
\multirow{-6}{*}{\rotatebox{90}{w. loop}}   & MambaVO++(Ours) & \textbf{0.012}  & 0.017                         & \textbf{0.020}  & \textbf{0.043} & \textbf{0.035} & \underline{0.035} & \textbf{0.008} & \underline{0.016} & \textbf{0.016} & 0.015                         & \underline{0.018} & \textbf{0.021} \\ \midrule
                            & DeepV2D\cite{teed2018deepv2d}         & 1.614                          & 1.492                         & 1.635                         & 1.775                         & \textbf{1.013} & 0.717                         & 0.695                         & 1.483                         & 0.839                         & 1.052                         & 0.591                         & 1.173                         \\
                            & TartanVO\cite{wang2021tartanvo}        & \underline{0.639}  & 0.325                         & 0.550                          & 1.153                         & \underline{1.021} & 0.447                         & 0.389                         & 0.622                         & \underline{0.433} & 0.749                         & 1.152                         & 0.680                          \\
                            & SVO\cite{forster2014svo}             & 0.10                            & 0.12                          & 0.41                          & 0.43                          & 0.30                           & 0.07                          & 0.21                          & \ding{55}                             & 0.11                          & 0.11                          & 1.08                          & -                             \\
                            & DROID-VO\cite{teed2021droid}        & 0.163                          & 0.121                         & 0.242                         & 0.399                         & 0.270                          & 0.103                         & 0.165                         & 0.158                         & 0.102                         & 0.115                         & \underline{0.204} & 0.186                         \\
                            & DPVO\cite{teed2024deep}            & 0.087                          & 0.055                         & \underline{0.158} & \textbf{0.137} & 0.114                         & 0.050                          & \underline{0.140}  & \textbf{0.086} & 0.057                         & \underline{0.049} & 0.211                         & \underline{0.105} \\
                            & V2V\cite{gurumurthy2024variance}             & 0.081                          & 0.067                         & 0.171                         & 0.179                         & 0.115                         & \underline{0.046} & 0.16                          & \underline{0.097} & 0.056                         & 0.059                         & 0.252                         & 0.117                         \\
\multirow{-7}{*}{\rotatebox{90}{w.o. loop}} & MambaVO(Ours)   & \textbf{0.063} & \textbf{0.024} & \textbf{0.107} & \underline{0.148} & 0.118                         & \textbf{0.044} & \textbf{0.138} & 0.1                           & \textbf{0.035} & \textbf{0.047} & \textbf{0.208} & \textbf{0.094} \\ \bottomrule
\end{tabular}}
\caption{ATE[m]$\downarrow$ results on the EuRoC. The top one is in \textbf{bold} and the second is \underline{underlined}.}
\label{tab:EuRoC}
\end{table*}

\begin{table*}[]
\centering
\scalebox{0.76}{
\begin{tabular}{c|cccccccccccc|c}
\toprule
                            & Method     & 00                              & 01                             & 02                             & 03                             & 04                            & 05                             & 06                             & 07                            & 08                              & 09        & 10                            & AVG                           \\ \midrule
                            & ORB-SLAM2\cite{mur2017orb}  & 8.27                           & \ding{55}                             & \textbf{26.86} & 1.21                          & \underline{0.77} & 7.91                          & 12.54                         & 3.44                         & \textbf{46.81}  & 76.54                         & \textbf{6.61}  & -                             \\
                            & ORB-SLAM3\cite{campos2021orb}  & \underline{6.77}   & \ding{55}                             & 30.50                          & \textbf{1.036} & 0.93                         & 5.542                         & 16.605                        & 9.70                          & \underline{60.69} & \textbf{7.89} & \underline{8.65}  & -                             \\
                            & LDSO\cite{gao2018ldso}       & 9.32                           & 11.68                         & 31.98                         & 2.85                          & 1.22                         & \underline{5.1}   & 13.55                         & 2.96                         & 129.02                         & \underline{21.64} & 17.36                         & \underline{22.42} \\
                            & DROID-SLAM\cite{teed2021droid} & 92.1                           & 344.6                         & \ding{55}                             & 2.38                          & 1.00                            & 118.5                         & 62.47                         & 21.78                        & 161.60                          & \ding{55}                             & 118.70                         & -                             \\
                            & DPV-SLAM\cite{lipson2024deep}   & 112.80                          & \underline{11.50}  & 123.53                        & 2.50                           & 0.81                         & 57.8                          & 54.86                         & 18.77                        & 110.49                         & 76.66                         & 13.65                         & 53.03                         \\
                            & DPV-SLAM++\cite{lipson2024deep} & 8.30                            & 11.86                         & 39.64                         & 2.50                           & 0.78                         & 5.74                          & \textbf{11.6}  & \textbf{1.52} & 110.9                          & 76.70                          & 13.70                          & 25.76                         \\
\multirow{-7}{*}{\rotatebox{90}{w. loop}}   & MambaVO++(Ours)  & \textbf{6.19}   & \textbf{8.04}  & \underline{27.73} & \underline{1.94}  & \textbf{0.59} & \textbf{3.05}  & \underline{11.79} & \underline{1.7}  & 105.42                         & 63.24                         & 10.51                         & \textbf{21.84} \\  \midrule
                            & DROID-VO\cite{teed2021droid}   & \textbf{98.43}  & 84.20                          & \underline{108.8} & 2.58                          & 0.93                         & 59.27                         & 64.4                          & 24.20                         & \textbf{64.55}  & \textbf{71.8}  & 16.91                         & 54.19                         \\
                            & DPVO\cite{teed2024deep}       & 113.21                         & \underline{12.69} & 123.4                         & \underline{2.09}  & \underline{0.68} & \underline{58.96} & \textbf{54.78} & 19.26                        & \underline{115.90}  & 75.10                          & \textbf{13.63} & \underline{53.03} \\
\multirow{-3}{*}{\rotatebox{90}{w.o. loop}} & MambaVO(Ours)    & \underline{112.39} & \textbf{8.16}  & \textbf{93.78} & \textbf{1.80}   & \textbf{0.66} & \textbf{56.51} & \underline{57.19} & \textbf{17.9} & 116.01                         & \underline{73.56} & \underline{14.37} & \textbf{50.21} \\ \bottomrule
\end{tabular}}
\caption{ATE[m]$\downarrow$ results on the KITTI dataset. The top one is in \textbf{bold} and the second is \underline{underlined}.}
\label{tab:kitti}
\end{table*}

\begin{table*}[h]
\centering
\scalebox{0.76}{
\begin{tabular}{c|cccccccccc|c}
\toprule
                            & Method          & 360    & desk                          & desk2                         & floor                         & plant                         & room                          & rpy                           & teddy  & xyz                           & AVG                           \\ \midrule
                            & ORB-SLAM2\cite{mur2017orb}       & \ding{55}                             & 0.071                         & \ding{55}                             & 0.023                         & \ding{55}                             & \ding{55}                             & \ding{55}                             & \ding{55}                             & 0.01                          & -                             \\
                            & ORB-SLAM3\cite{campos2021orb}       & \ding{55}                             & \underline{0.017} & \textbf{0.21}  & \ding{55}                             & 0.034                         & \ding{55}                             & \ding{55}                             & \ding{55}                            & \textbf{0.009} & -                             \\
                            & GO-SLAM\cite{zhang2023go}         & \underline{0.089} & \textbf{0.016} & \underline{0.028} & \underline{0.025} & 0.026                         & \underline{0.052} & \textbf{0.019} & \underline{0.048} & \underline{0.01}  & \underline{0.035} \\
                            & DROID-SLAM\cite{teed2021droid}      & 0.111                         & 0.018                         & 0.042                         & \textbf{0.021} & \textbf{0.016} & \textbf{0.049} & \underline{0.026} & \underline{0.048} & 0.012                         & 0.038                         \\
                            & DPV-SLAM\cite{lipson2024deep}        & 0.112                         & 0.018                         & 0.029                         & 0.057                         & \underline{0.021} & 0.33                          & 0.03                          & 0.084                         & \underline{0.01}  & 0.076                         \\
\multirow{-6}{*}{\rotatebox{90}{w. loop}}   & MambaVO++(Ours) & \textbf{0.085} & \textbf{0.016} & 0.032                         & 0.027                         & 0.025                         & 0.056                         & \underline{0.026} & \textbf{0.029} & \textbf{0.009} & \textbf{0.034} \\ \midrule
                            & TartanVO\cite{wang2021tartanvo}        & 0.178                         & 0.125                         & 0.122                         & \underline{0.349} & 0.297                         & \textbf{0.333} & 0.049                         & 0.339                         & 0.062                         & 0.206                         \\
                            & DeepV2D\cite{teed2018deepv2d}         & 0.182                         & 0.652                         & 0.633                         & 0.579                         & 0.582                         & 0.776                         & 0.053                         & 0.602                         & 0.15                          & 0.468                         \\
                            & DeepFactors\cite{czarnowski2020deepfactors}     & 0.159                         & 0.17                          & 0.253                         & 0.169                         & 0.305                         & \underline{0.364} & 0.043                         & 0.601                         & 0.035                         & 0.233                         \\
                            & DPVO\cite{teed2024deep}            & \underline{0.135} & 0.038                         & 0.048                         & 0.04                          & 0.036                         & 0.394                         & \underline{0.034} & 0.064                         & \textbf{0.012} & \underline{0.089} \\
                            & V2V\cite{gurumurthy2024variance}             & 0.145                         & \underline{0.026} & \underline{0.044} & 0.064                         & \underline{0.031} & 0.434                         & 0.045                         & \textbf{0.046} & \textbf{0.012} & 0.094                         \\
\multirow{-6}{*}{\rotatebox{90}{w.o. loop}} & MambaVO(Ours)   & \textbf{0.108} & \textbf{0.021} & \textbf{0.037} & \textbf{0.034} & \textbf{0.022} & 0.372                         & \textbf{0.031} & \underline{0.048} & \underline{0.013} & \textbf{0.076} \\ \bottomrule
\end{tabular}}
\caption{ATE[m]$\downarrow$ results on the TUM-RGBD dataset. We only conduct experiments under the monocular setting. The top one is in \textbf{bold} and the second is \underline{underlined}. }
\label{tab:tum}
\end{table*}

\subsection{Analysis on Matching Improvement}
\label{sec:matching}
\vspace{-0.1in}
\begin{table}[H]
\centering
\scalebox{0.77}{
\begin{tabular}{c|cccc}
\toprule
Method         & AUC@1°$\uparrow$ & AUC@2°$\uparrow$ & AUC@5°$\uparrow$ & AUC@10°$\uparrow$ \\ \midrule
DROID-VO\cite{teed2021droid}      & 0.123  & 0.167  & 0.294  & 0.53    \\
DPVO\cite{teed2024deep}           & 0.299  & 0.485  & 0.738  & 0.925   \\
V2V\cite{gurumurthy2024variance}  & \underline{0.399}  & \underline{0.584}  & \underline{0.855}  & \underline{0.957}   \\
MambaVO (Ours)                    & \textbf{0.471}  & \textbf{0.678}  & \textbf{0.892}  & \textbf{0.975}   \\ \bottomrule
\end{tabular}}
\caption{Geometric matching experiments on EuRoC dataset. The top one is in \textbf{bold} and the second is \underline{underlined}.}
\label{tab:matching}
\end{table}

To further evaluate how the Mamba architecture contributes to the matching improvement, we evaluate the image matching metric separately. The compared methods include other visual odometry algorithms. Disabling bundle adjustment and optimization, we use only the matching outputs to compute relative poses between frames via Poselib \cite{PoseLib}, which employs epipolar error \cite{zhang1998determining} and LO-RANSAC \cite{chum2003locally}.
Poses are evaluated using the AUC (Area Under the Curve) metric, following the image matching methods in \cite{cai2024prism,wang2024efficient}.

As in \cref{tab:matching}, MambaVO achieves the best matching performance compared with previous methods on EuRoc dataset. 
DROID-VO relying on the dense flow matching, often introduces noise in correspondences in challenging areas.
DPVO and V2V samples pixel patches and perform matching by inner product on image feature, resulting in poor fine-matching capability.
Our GMM enhances pixel-level matching accuracy, especially improving AUC@1° and AUC@2° by 17\% and 16\%, respectively, over the previous best.
This highlights the effectiveness of our geometric initialization and Mamba-based sequential refinement.

\subsection{Ablation Study}
We conduct ablation studies to study the impacts of different components and different design choices. 
\subsubsection{Impact of Feature Selection in GIM}
\vspace{-0.1in}
\begin{table}[H]
\centering
\scalebox{0.8}{
\begin{tabular}{c|ccccc}
\toprule
Feature Selection & SIFT\cite{lindeberg2012scale}  & ORB\cite{rublee2011orb}   & SP\cite{detone2018superpoint} & Patch\cite{teed2024deep} & Ours \\ \midrule
TartanAir         & 4.29  & 5.07  & 2.35       & \underline{0.18}         & \textbf{0.14}              \\
EuRoC             & 0.524 & 1.184 & 1.758      & \underline{0.115}        & \textbf{0.094}             \\ \bottomrule
\end{tabular}}
\caption{Ablation experiments on matching feature. We report the ATE[m]$\downarrow$ results on EuRoC dataset. The top one is in \textbf{bold} and the second is \underline{underlined}.}
\label{tab:feature}
\end{table}
In GIM, we used the geometric feature of EfficientLoFTR, i.e. $\mathbf G_t$ for matching.  
We replace the feature matching by 
SIFT \cite{lindeberg2012scale}, ORB \cite{rublee2011orb} , Superpoint \cite{detone2018superpoint}, and random patches \cite{teed2024deep} respectively to perform ablation experiments on the EuRoC dataset to figure out how the matching feature affects the deep visual odometry. The results are shown in \cref{tab:feature}.
We demonstrate the obvious ATE decreasing in using our semi-dense feature.  
\subsubsection{Impact of  Modules in GIM, GMM and TAP}
\begin{table}[h]
\centering
\scalebox{0.8}{
\begin{tabular}{cccc}
\toprule
\multicolumn{2}{c}{Configuration}        & EuRoC & KITTI \\ \midrule
\multirow{3}{*}{GIM} & w.o. Context feature   & 0.181 & 58.62  \\
                     & w.o. Geometric feature & 0.309 & 192.75  \\
                     & w.o. PnP               & 0.199 & 201.16  \\ \midrule
\multirow{3}{*}{GMM} & w.o. History fusion    & 0.179 & 59.97  \\
                     & w.o. Mamba blocks      & 0.318 & \ding{55}      \\
                     & w.o. GRU               & 0.293 & 190.01 \\ \midrule
\multirow{3}{*}{TAP} & w.o. Gradient weight   & 0.111 & 50.87  \\ 
                     & w.o. History balance   & 0.101 & 51.93  \\ \midrule
\multicolumn{2}{c}{MambaVO (full)}       & \textbf{0.094} & \textbf{50.21} \\ \bottomrule
\end{tabular}}
\caption{Ablation experiments on modules in MambaVO. We report the ATE[m]$\downarrow$ results. The top one is in \textbf{bold}. 
}
\label{tab:ablation}
\end{table}

The ablation study results presented in \cref{tab:ablation} provide the impacts of components in the GIM, GMM, and TAP. The experiments are conducted on the EuRoC and KITTI and we report the ATE metric.
When evaluating GIM's components, we fix the GMM and TAP,  and vice visa.  

The results indicate that in GIM,  both the context features and geometric features significantly contribute to reducing ATE.  PnP is crucial for accuracy,  especially on the KITTI outdoor dataset, where the localization error rises significantly without the initial poses of PnP.
For GMM, removing any key component degrades performance. In particular, removing the Mamba blocks causes the failure on KITTI. 
In addition, our TAP strategy during training contributes to the accuracy in unseen scenarios like EuRoC and KITTI.
The full MambaVO achieves the best performance with all the components.
\subsubsection{Impact of Smoothed Training Loss}
\begin{figure}[ht]
  \centering
  \includegraphics[width=0.81\linewidth]{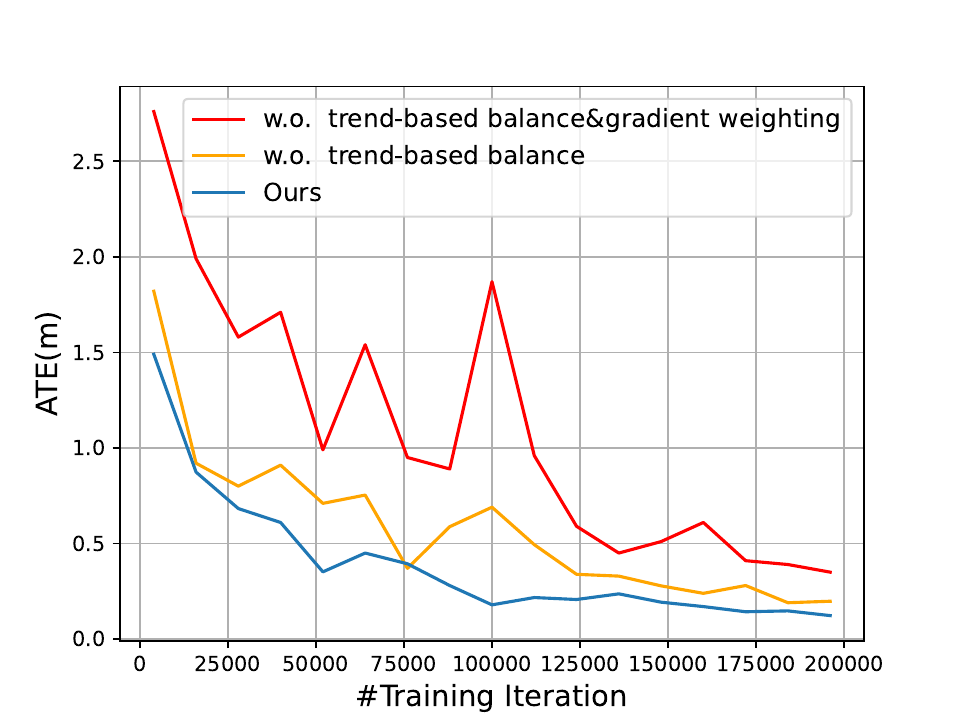}
  \caption{We report the average ATE on the validation split of TartanAir. We observe that our loss design strategy makes training converge faster and achieve smaller ATE error.}
  \label{fig:loss}
\end{figure}

We evaluate the two parameters for training loss smoothing (\cref{sec:smoothing}), one is the gradient weighting parameter and the other is the trend-based balance parameter. 
We retrained MambaVO without the either kind of the parameter and report the ATE on the TartanAir validation set. The results are shown in \cref{fig:loss}. 
We can observe that with the assistance of these two parameters, the loss curve converges faster and more stably. It also achieves lower ARE on the validation set under the same training iterations. 

\vspace{-0.1in}

\subsection{Time and Memory}
\label{sec:time}

MambaVO can run in real-time on a single RTX 3090 GPU. 
On EuRoC, we achieve an average real-time performance of 22Hz, exceeding the camera frame rate of 20Hz. On the TUM-RGBD and KITTI datasets, the frame rates are 30Hz and 23Hz, respectively. All of the results are obtained under the settings shown in the \cref{tab:EuRoC}, \cref{tab:kitti}, \cref{tab:tum}.

\begin{figure}[htbp]
  \centering
  \includegraphics[width=0.9\linewidth]{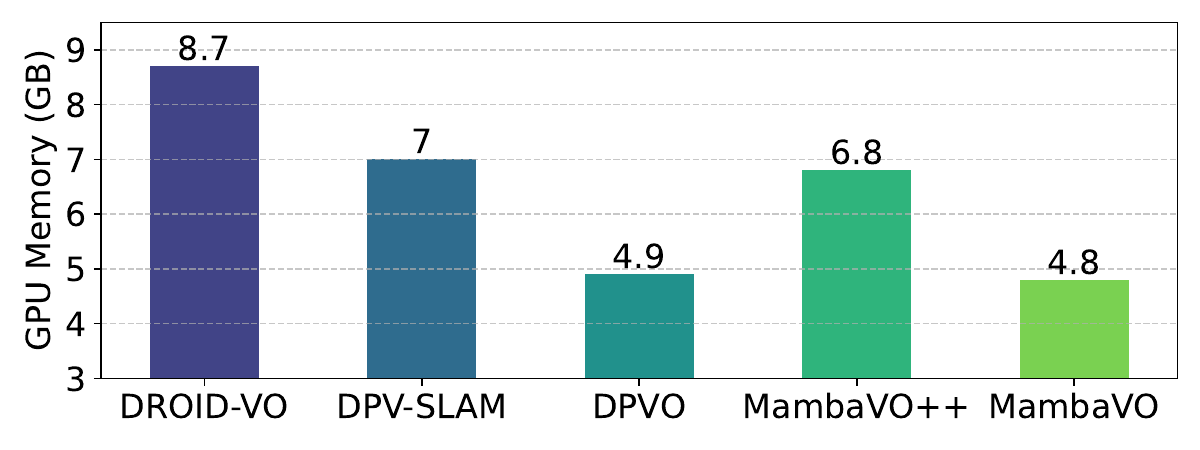}
  \caption{GPU memory usage of MambaVO and MambaVO++ compared to other methods. }
  \label{fig:memory}
  \vspace{-0.2in}
\end{figure}

We also compared GPU memory usage for inference with current state-of-the-art deep visual odometry methods. Experimental results demonstrate that our method has smaller GPU memory consumption during inference, as the results shown in \cref{fig:memory}.
\section{Conclusion}
This paper proposes MambaVO, a novel deep visual odometry system to address the accuracy and robustness of deep VO based on learning to optimize.
We conduct reliable initialization,  sequential matching refinement, and training smoothing based on the Mamba architecture.  
Our method achieves the state-of-the-art results on TartanAir, EuRoC, TUM-RGBD, and KITTI. In the future, we will utilize dense information and 3D Gaussian Splatting \cite{kerbl20233d} technology to obtain high-fidelity maps for rich scene representation while aintaining localization accuracy.

{
    \small
    \bibliographystyle{ieeenat_fullname}
    \bibliography{main}
}



\end{document}